\documentclass[runningheads]{llncs}

\usepackage{eccv}

\usepackage{eccvabbrv}

\usepackage{graphicx}
\usepackage{booktabs}

\usepackage[accsupp]{axessibility}  

\usepackage{hyperref}
\usepackage{wrapfig}
\usepackage{orcidlink}
\usepackage{amsmath}
\usepackage{multirow}
\begin{document}

\title{Self-supervised Shape Completion via Involution and Implicit Correspondences} 

\titlerunning{Ssp.~Shape Completion via Involution and Correspondences}

\author{Mengya Liu\inst{1}\orcidlink{0009-0006-9122-6698} 
\and
Ajad Chhatkuli\inst{1, 3}\orcidlink{0000-0003-2051-2209} 
\and
Janis Postels\inst{1}\orcidlink{0000-0002-3490-1726} \and \\
Luc Van Gool\inst{1, 3}\orcidlink{0000-0002-3445-5711} 
\and Federico Tombari\inst{2}\orcidlink{0000-0001-5598-5212}}

\authorrunning{M.~Liu, A.~Chhatkuli et al.}

\institute{{Computer Vision Lab, ETH Zurich} \and  
{Google, TU Munich} \and {INSAIT, Sofia University} \\
\scriptsize{\email{\{mengya.liu, ajad.chhatkuli, jpostels, vangool\}@vision.ee.ethz.ch, tombari@google.com}}}

\maketitle
\begin{abstract}
3D shape completion is traditionally solved using supervised training or by distribution learning on complete shape examples. Recently self-supervised learning approaches that do not require any complete 3D shape examples have gained more interests. In this paper, we propose a non-adversarial self-supervised approach for the shape completion task. Our first finding is that completion problems can be formulated as an involutory function trivially, which implies a special constraint on the completion function $\mathcal{G}$, such that $\mathcal{G} \circ \mathcal{G}(X) = X$. Our second constraint on self-supervised shape completion relies on the fact that shape completion becomes easier to solve with correspondences and similarly, completion can simplify the correspondences problem. We formulate a consistency measure in the canonical space in order to supervise the completion function. We efficiently optimize the completion and correspondence modules using ``freeze and alternate'' strategy.
The overall approach performs well for rigid shapes in a category as well as dynamic non-rigid shapes. We ablate our design choices and compare our solution against state-of-the-art methods, showing remarkable accuracy approaching supervised accuracy in some cases.
\end{abstract}
\section{Introduction}
Shape completion for partial shape scans of a category has only recently been tackled as a self-supervised learning task~\cite{cui2023p2c,kim2023learning,wen2021cycle4completion,chen2019unpaired,peters2022self}, among which some~\cite{wen2021cycle4completion,chen2019unpaired} use a training set of complete examples. Despite not having complete shape examples due to self-occlusions, human vision can reliably understand complete shapes of objects~\cite{sekuler2000visual}, possibly due to recognition and shape understanding. Our goal is to explore a similar ability of explicit low-level shape understanding in a neural network for shape completion.

State-of-the-art approaches for shape completion either rely on full supervision~\cite{yuan2018pcn, zhou2022seedformer, yan2022shapeformer}
or on adversarial learning through unpaired examples of partial and complete shapes~\cite{wen2021cycle4completion, chen2019unpaired, wu2020multimodal}. The challenges of learning shape completion without complete examples are significant, as this requires a thorough exploration of the relationship between different 3D shapes within a given category. Recent work \cite{peters2022self} explores the distribution of local surface patches in order to devise an adversarial learning problem without complete shape examples. Even more recently \cite{cui2023p2c} explores local similarities between shape patches in order to devise self-supervised shape completion. We argue that local patch distributions may be insufficient for accurate completion and explore novel directions for the task.

In this paper, we study the problem of self-supervised partial shape completion via correspondence estimation. Similar to the state-of-the-art methods~\cite{cui2023p2c,kim2023learning,wen2021cycle4completion,chen2019unpaired,peters2022self}, we perform completion of rigidly aligned partial shapes. In order to use correspondences for completion, we parameterize correspondences via a deformation warp from the input shape into a common template or canonical space~\cite{zheng2021deep,deng2021deformed,lei2022cadex}. Such canonical pasteurization of shapes in a category allows shape reconstruction as a function of correspondences. We base the network architecture on the hypothesis that better completion can induce better correspondences and vice-versa. The quality of the correspondences in this case can be measured by the deformation consistency of many shapes onto the common template space~\cite{lei2022cadex,liu2023unsupervised}. In other words, completion can be guided by the objective of having better consistency of shape deformations to a common space. We supervise the deformation warp using regularization and reconstruction accuracy similar to \cite{zheng2021deep}. Our completion module uses a point generator (the completion function) and an upsampler~\cite{xiang2021snowflakenet}. Moreover, we observe that completion functions, regardless of their domains, can be formulated to be an exact involution. An involution is a function which is an inverse of itself. In contrast to e.g., standard cyclic constraints, involutions are highly restrictive constraints but rarely applicable. Consequently, involution allows us to devise a strong self-supervised loss without using any external augmentation or external data prior.  We propose representing both incomplete and complete shapes using Implicit Neural Representation~\cite{zheng2021deep,park2019deepsdf, mescheder2019occupancy}. As partial shapes are often non-watertight, we use an Unsigned Distance Field (UDF) representation~\cite{chibane2020neural,guillard2022meshudf,zhou2022learning, zhou2023learning}.

Despite the simple design, the joint optimization of completion and correspondence estimation is ill-posed. Taking inspiration from Generative Adversarial Networks (GANs), we opt for a batch-wise alternate optimization of the losses, where each iteration optimizes only one of the two sets of parameters for completion and correspondences, leading to the desired convergence. 
We evaluate and ablate our methodology using partial shapes of both synthetic data and real scans. Results show that our method performs superior to other self-supervised methods and the unpaired adversarial approaches. Overall, our main contributions can be summarized as:
\begin{enumerate}
    \item We propose to formulate the completion task as an involution, which implies a special property of the completion function $\mathcal{G}$, such that $\mathcal{G} \circ \mathcal{G}(X') = X'$ where $X'$ is the partial point set input.
    \item We supervise completion using the consistency of the canonical space, which is a proxy measure for correspondence quality. We supervise correspondences using the partial shape reconstruction loss and deformation regularizers. We show these losses are enough for the self-supervision of shape completion.
    \item We use a batch-wise training strategy to alternately optimize the completion module and the template-based INR module.
    \item We provide a partial shape UDF dataset that contains the dense partial surface and the UDF field for each instance.
\end{enumerate}

\section{Related works}
\textbf{Supervised Shape Completion.}
Shape completion is largely tackled as a supervised learning problem with paired data~\cite{xiang2021snowflakenet, zhou2022seedformer, huang2020pf, liu2020morphing, wen2020point, dai2017shape, yan2022shapeformer, rao2022patchcomplete, chu2023diffcomplete, mo2019partnet, xie2020grnet, yuan2018pcn}. 3D-EPN~\cite{dai2017shape} trains an autoencoder to complete the shape in voxelized representation. PatchComplete~\cite{rao2022patchcomplete} uses the multi-resolution local patch prior and completes the SDF grid. DiffComplete~\cite{chu2023diffcomplete} learns a diffusion model conditioned on the incomplete shape and outputs the occupancy representation. In favor of fine-grained details, \cite{xiang2021snowflakenet, zhou2022seedformer, zhang2020detail, wen2021pmp} adopt a coarse-to-fine strategy. They first generate the coarse points, and upsample them step by step. Additionally, the use of self-attention via transformers~\cite{parmar2018image, zhao2021point, yan2022shapeformer,yu2021pointr, zhou2022seedformer, xiang2021snowflakenet} show improved performance.
Snowflakenet~\cite{xiang2021snowflakenet} progressively upsamples points
where the child points are recursively generated from parent points using a skip-transformer. 
Recently \cite{zhou2022seedformer} introduced the patch seeds and an upsample transformer for better local details.

\noindent\textbf{Unpaired Shape Completion.}
Although supervised approaches show their strong capability in shape completion, they are limited by the availability of paired data, for example in real scans. Such paired data may not be available in all categories. Consequently, many works take advantage of unpaired examples of complete and partial shapes. cGan~\cite{chen2019unpaired} first proposes to use the unpaired data to achieve shape completion through adversarial loss. However, it needs to first train two auto-encoders with the complete shape examples and partial shapes respectively. The follow-up work ~\cite{wu2020multimodal} created a variational autoencoder to learn multiple shapes with one partial input. Later many works~\cite{zhang2021unsupervised, wen2021cycle4completion, cao2021mfm} have designed the partial-to-full latent code transformation using the adversarial loss. ShapeInversion~\cite{zhang2021unsupervised} proposes GAN inversion to learn the shape removal to go from full shape to partial shape with a pre-trained GAN from the full shape. The latest DDIT~\cite{li2023ddit} pre-trains a deep implicit template that holds the deformation from various shapes to the template shape. Later the shapes are completed by deforming the template with the latent code conditioned on the partial shapes. SCoDA~\cite{wu2023scoda} at the same time, solves the domain gap between synthetic data completion and real data by transferring knowledge with the cross-domain feature fusion method.

\noindent\textbf{Self-supervised Shape Completion.}
There exist very few truly self-supervised shape completion approaches. They do so using the local patch or local shape priors~\cite{chu2021unsupervised, kim2023learning, peters2022self, cui2023p2c}. \cite{chu2021unsupervised} uses test-time optimization to directly learn contextual information from the known regions to complete a shape based on the neural tangent kernel. \cite{kim2023learning} on the other hand, takes advantage of the poses from multi-views to integrate the complete point cloud. \cite{cui2023p2c} generates patches of the partial point cloud in order to achieve both shape augmentation and region-aware regularization. \cite{peters2022self} develop an adversarial framework by using a discriminator network to reject incomplete shapes via a loss function that separately assesses local sub-regions of the generated example and accepts only regions with sufficiently high point count in a voxelized representation. On the other hand, we solve the task of shape completion by observing similarities between completed shapes in a category. Similar to DDIT~\cite{li2023ddit}, but without a pre-trained template shape, our method learns to generate a consistent template space gradually from only the partial inputs and generates the dense meshes.
\section{Background and Method Architectures}
\label{sec:prelim}
\begin{figure*}[t]
    \centering
    \includegraphics[width=1.0\textwidth]{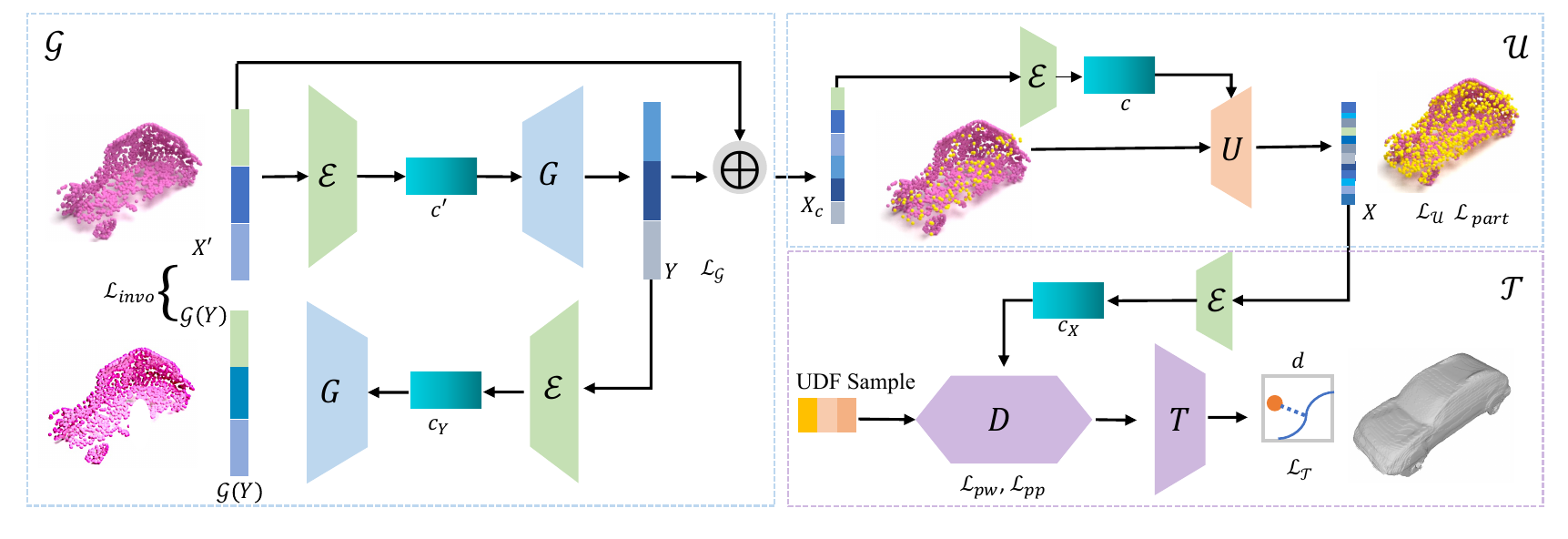}
    \caption{Architecture of our method. We show the detailed implementation of the completion module ($\mathcal{G}$ and $\mathcal{U}$), and the template-based INR module $\mathcal{T}$. The completion function $\mathcal{G}$ starts with partial input $X'$, where an encoder $\mathcal{E}$ is used to extract a shape code $c'$ and is fed to the decoder $G$ to produce $Y$. The involution is implemented by feeding $Y$ to $\mathcal{G}$ that should ideally result in the input partial points. The concatenated coarse but complete points $X_c$ go through upsampler $\mathcal{U}$ and generates detailed shape points $X$ and provides conditioning on $\mathcal{T}$. $\mathcal{L}$ denotes the losses applied on each module. }
    \label{fig:completion_modules}
\end{figure*}

In this section, we outline our architectural design for self-supervised shape completion and correspondences. We illustrate the architecture in Figure \ref{fig:completion_modules}.

\subsection{Overview}
The method consists of two modules, the completion module and the template-based INR module. The completion module consists of a completion function (or a generator) $\mathcal{G}$ and an umsampler function $\mathcal{U}$. We first generate the missing part using $\mathcal{G}$ and get the detailed shape after the upsampler $\mathcal{U}$. Later we feed all points to the template based INR $\mathcal{T}$ in order to obtain the completed shape implicit field. The two modules are supervising each other during training. During the inference, we obtain the final completed shape using Marching Cubes (MC)~\cite{guillard2022meshudf} on the final implicit field. Next, we describe each module in necessary details.

\subsection{Completion Module ($\mathcal{G}$ and $\mathcal{U}$)}
In contrast to unsupervised shape correspondences, where multiple paradigms exist for solving the problem (including ones with template-based  INR~\cite{zheng2021deep,lei2022cadex}), self-supervised shape completion is rather ill-posed and has a very few classes of solutions. Furthermore, we want to devise a completion module that does not need full shape examples. Nonetheless, we re-purpose the completion module proposed in Snowflakenet~\cite{xiang2021snowflakenet} to predict sample points on the missing parts of the shape guided by self-supervised losses that will be described in Section~\ref{sec:unsup}.

Formally, let $X'$ denote the set of points on the partial shape $\mathcal{X'}$. Then the completion function $\mathcal{G}$ takes as input $X'$ and outputs the set of points $Y$ which contain missing points, i.e., $\mathcal{G}(X') = Y$. The concatenation of the points $Y$ and the initial points $X'$ thus provides the complete coarse shape points $X_c$. As we will see further on, this definition of the completion function provides a natural self-supervision loss. The architecture of the completion function $\mathcal{G}$ consists of an encoder $\mathcal{E}$ providing a single feature vector and a decoder $G$ resulting in the missing points.
\begin{equation}
    \mathcal{G}(X';\theta_{\mathcal{G}}) = G \circ \mathcal{E} = Y
\end{equation}
$\theta_\mathcal{G}$ denotes the network parameters of the completion function. Our completion module has an additional purpose: the encoder also provides a latent code describing the given set of points. Mathematically,
\begin{equation}
\label{eq:encoder}
\mathcal{E}(X') = c', \quad \mathcal{E}(Y) = c_Y \quad \text{and} \quad \mathcal{E}(X_c) = c.
\end{equation}
The latent codes provide additional context for the modules in the forward path.

The upsampler $\mathcal{U}$ obtains finer shape by increasing the number of points by splitting each parent point into multiple child points. We refer the reader to \cite{xiang2021snowflakenet} for detailed model architecture. Thus, the upsampler consists of the encoder $\mathcal{E}$ and the upsampler decoder $U$ to get the detailed shape.
\begin{equation}
        \mathcal{U}(X_c;\theta_{\mathcal{U}}) = U \circ \mathcal{E} = X \\
\end{equation}
Here, $X$ denotes the final set of points from the completion module. $U$ is the decoder that provides child points from $X_c$ guided by the shape code $c$ extracted from the encoder $\mathcal{E}$ (see in Eq.~\eqref{eq:encoder}). $\theta_\mathcal{U}$ represents the network parameters of the upsampler.
The upsampled points further provide better latent code to condition the warp function $D$ in the template-based INR module $\mathcal{T}$.

\subsection{Template-based INR Module $\mathcal{T}$} 
As partial shapes cannot be represented using the Signed Distance Fields (SDF), we sample them using the Unsigned Distance Fields (UDF). Instead of directly fitting an INR on the partial shape similar to \cite{park2019deepsdf}, we use template-based INR~\cite{deng2021deformed,zheng2021deep,lei2022cadex,li2023ddit, liu2023unsupervised} as it encodes the shape correspondences in a category. At any point $x\in\mathbb{R}^3$ that is around the shape surface,
the INR can be written as:
\begin{equation}
\label{eq:inrcomp}
    \mathcal{T}(x; c_X,\theta_\mathcal{T}) = T\circ D = d.
\end{equation}
We collectively denote the network parameters as $\theta_\mathcal{T}$ for the Template INR $\mathcal{T}$. Essentially, the warp $D$ transforms an input point $x$ near the shape surface to a consistent template space before passing it to the INR decoder $T$. Thus, $D(x;c_X)$ is a function of a point in space conditioned on the latent code of the shape instance $c_X$ extracted with the encoder $\mathcal{E}$. Finally $d$ denotes the UDF distance for the point $x$ to its closest surface point.

State-of-the-art methods train the template-based INR with complete shape instances of a category, using the reconstruction loss on the output of $T$. In such cases, Eq.~\eqref{eq:inrcomp} can lead to a consistent template space, which thus results in valid correspondences across the input shapes~\cite{zheng2021deep,deng2021deformed,lei2022cadex, liu2023unsupervised}. In case of only partial shape examples, the warp $D$ cannot deform to a consistent template. Therefore, we provide the completed (although naturally with errors) points $X$ to the warp $D$. In the next section, we describe the training of the completion ($\theta_\mathcal{G}$), upsampler ($\theta_\mathcal{U}$) and the template-based INR ($\theta_\mathcal{T}$) parameters.

\section{Self-supervised Constraints and Training}
\label{sec:unsup}
The self-supervised training of the completion module and the template-based INR module on incomplete shapes form the core of our approach. We introduce the proposed constraints and describe their use in training as follows.

\subsection{Completion Self-constraint}
\label{sec:self-constrain}
The definition of the completion function $\mathcal{G}(X')=Y$ means that any partial shape points $X'$ as input results in the missing points $Y$. Conversely, if $Y$ is used as the initial points, by definition, the completion function $\mathcal{G}(Y)$ should result in the points $X'$. Thus, the completion function $\mathcal{G}$ by definition is an exact \emph{involution}. Consequently, the ideal completion function must satisfy the following constraint:
\begin{equation}
\label{eq:generator}
    \mathcal{G} \circ \mathcal{G} (X') = \mathcal{G} (Y) = X'.
\end{equation}
Although Eq.~\eqref{eq:generator} is defined for shape completion, it is straight-forward to note that Eq.~\eqref{eq:generator} is true for any local representation of $X'$ (\eg, point clouds, pixels, voxels in contrast to non-local \eg SDF or UDF).
We represent point sets as matrices for implementation, but in fact Eq.~\eqref{eq:generator} are equalities of sets. Thus we use the following loss on $\mathcal{G}$:
\begin{equation}
\label{eq:genloss}
    \mathcal{L}_{invo} = \operatorname{Chamfer}(\mathcal{G}(Y) - X').
\end{equation}
Here $\operatorname{Chamfer}(.)$ is the operator for the bi-directional Chamfer distance. Eq.~\eqref{eq:genloss} is the self-loss of the completion function, derived from its definition. Note that the identity function and $\mathcal{G}(X) = 1/X$ are trivial solutions for involution. However, using further constraints, specifically, the template shape consistency loss, we guide the optimization away from the trivial solutions. Nonetheless, the involution restrains the function $\mathcal{G}$ to be a completion function.

\textbf{Involution and the cyclic constraints.}
Cyclic constraints \cite{kalal2010forward,ginzburg2020cyclic,wen2021cycle4completion,zhu2017unpaired} in computer vision problems have been known for their efficacy as well as elegance. They are also associated to a fundamental conservation principle in Kirchoff's second (voltage) law~\cite{feynman1963feynman} and in transport or graph flow problems~\cite{masch1980cyclic}. A cyclic constraint is defined by two different transformations/functions $f,g$ such that $g \circ f (x) = x$. Thus, the involution constraint in Eq.~\eqref{eq:generator} is a special cyclic constraint with $g = f$, which is a stricter condition. Although involution itself has many interesting properties, for the purpose of this work, it is only important to note that the solution space of $f = f^{-1}$ is much smaller than that of $f = g^{-1}$. One more note on involution in machine learning is that it can be interpreted as self-augmentation. The second use of the function $f$ in $f(f(x))$ is thus on the augmented data obtained via prediction $f(x)$.

\subsection{Completion Module Supervision Loss}
As highlighted in Section~\ref{sec:prelim}, the template INR \emph{requires} complete shape examples in a category for obtaining consistent template space. This essentially provides a key supervisory constraint for us to optimize the completion module. We thus use the template space consistency similar to \cite{liu2023unsupervised} but in order to supervise the completion module rather than the warp. We first generate the template shape $P$ by directly predicting the zero level set in the warp space, i.e., by taking the zero level set of $T(x)$. For more details, we refer the reader to \cite{zheng2021deep}. Thus the template consistency loss is designed to constrain each warped shape (with generated points) to be close to the template shape $P$. Given $X$ as the shape points that are output from the completion module, we obtain the following loss for the completion module parameters $\theta_\mathcal{G}, \theta_{\mathcal{U}}$:
\begin{equation}
    \label{eq:loss_consistency}
    \mathcal{L}_{\mathcal{G}} = \operatorname{Chamfer}(\mathcal{D}(X_c; c), P),\   \mathcal{L}_{\mathcal{U}} = \operatorname{Chamfer}(\mathcal{D}(X; c_X), P)
\end{equation}
Here, $\operatorname{Chamfer}(.,.)$ is the operator that measures the bidirectional Chamfer distance between two sets of points. Moreover, we follow \cite{xiang2021snowflakenet} to use a partial matching loss $\mathcal{L}_{part}$ which forces the output point cloud $X$ from the upsampler $\mathcal{U}$ to partially match the input $X_c$. Therefore, we use a single-side Chamfer Distance to constrain the loss.
\begin{equation}
    \label{eq:partial_loss}
    \mathcal{L}_{part} = \underset{X_c\rightarrow X}{\operatorname{Chamfer}}(X_c, X)
\end{equation}
We use the following loss in order to optimize the completion network:
\begin{equation}
    \label{eq:loss_compl}
    \begin{aligned}
    \mathcal{L}_{\theta_\mathcal{G}} &= \mathcal{L}_\mathcal{G} + l_1 \mathcal{L}_{invo} \\
    \mathcal{L}_{\theta_\mathcal{U}} &= \mathcal{L}_\mathcal{U} + l_2 \mathcal{L}_{part} \\
    \end{aligned}
\end{equation}
Here, $l_1=1$ and $l_2=1$ for optimizing the completion module using a mini-batch.

\subsection{Template-based INR Module Loss}
While the ground-truth complete shapes are unknown for any shape instance, the partial shape is always available for supervising the INR $\mathcal{T}$. We thus use the partial shape reconstruction loss in order to guide the INR. Collectively we can define the constraint as follows:
\begin{equation}
    \label{eq:loss_inr}
    \mathcal{L}_{\mathcal{T}} = \left|\mathcal{T}([X]; c_X) - [d]\right|.
\end{equation}
$[X]$ denotes the set of neighboring points in matrix form with the UDF distances $[d]$ from the initial points $X'$ on the partial shape surface, as well as the generated points, where the UDF are assigned to be 0. The loss formulates the base of the template implicit field. The operator $[\ ]$ emphasizes the matrix/vector representation.

Moreover, it is clear that the deformation warp $D$ is central to the optimization of INR as well as the completion module. Thus, with strong non-linearities in $D$, the overall solution may be pushed towards incorrect correspondences and completion despite the other losses. In order to counteract that, similar to \cite{zheng2021deep,liu2023unsupervised} we use an identity regularization $\mathcal{L}_{pw}$ and a smoothness regularization $\mathcal{L}_{pp}$ as:
\begin{equation}
    \begin{aligned}
    \mathcal{L}_{pw} &= \sum_{{x} \in [X']}h({\lVert D({x}; c_X) - \mathsf{x}\rVert}_2)\\
    \mathcal{L}_{pp} &=  \sum_{{x},{y} \in [X']} \sum_{{x} \not={y}} \max(\frac{||\triangle {x} - \triangle {y}||_2}{||{x}-{y}||_2}, 0)
    \end{aligned}
\end{equation}
Here, $h(.)$ is the Huber function and $\triangle {x} = D({x};c_X) - \mathsf{x}$ is the position shift of a point ${x} \in [X']$. The loss $\mathcal{L}_{pw}$ prevents excessive deviation of a point, and $\mathcal{L}_{pp}$ minimizes the relative space distortions. We use the following loss for the INR module.
\begin{equation}
    \label{eq:loss_inr}
    \mathcal{L}_{\theta_\mathcal{T}} = \mathcal{L}_\mathcal{T} + l_3 \mathcal{L}_{pw} + l_4 \mathcal{L}_{pp}.
\end{equation}
We choose $l_3=0.0005, l_4=0.0001$. 

\subsection{Batch-wise Alternate Training}
Now that we have defined the complete set of losses, we describe how we use them in order to alternately optimize the parameters $\theta_\mathcal{T}$, $\theta_\mathcal{G}$ and $\theta_{\mathcal{U}}$. 
We observe that optimizing both parameters jointly does not result in the desired convergence. It is ill-posed to optimize both modules in one batch. Taking inspiration from GANs, we use the batch-wise alternate training strategy. To be specific, in the first batch, we train the template INR module with its losses while freezing the completion module parameters. Later we optimize the completion module parameters while freezing the template INR module. We exchange the two modules training alternatively, until the training converges.

We optimize the losses $\mathcal{L}_{\theta_\mathcal{T}}$
and $\mathcal{L}_{\theta_{\mathcal{G}}, \theta_\mathcal{U}}$ for a total of 50K iterations alternately. More precisely, we freeze the parameters $\theta_{\mathcal{G}}, \theta_\mathcal{U}$ while optimizing $\theta_\mathcal{T}$ in a mini-batch and vice-versa.

\section{Experimental Results}
\begin{table}[t]
\fontsize{8}{10}\selectfont
\setlength{\tabcolsep}{0.55mm}
    \centering
    \begin{tabular}{l |c| c c c c c c | c c c c c c }
    \toprule
        & &  \multicolumn{6}{c|}{F1} & \multicolumn{6}{c}{CD} \\

          & type & car & plane & sofa & chair & table & avg. & car & plane & sofa & chair & table & avg.\\
         \midrule
         SeedF~\cite{zhou2022seedformer} & sp. & 80.85 & 97.08 & 71.30 & 80.65 & 83.82 & 82.74 & 0.82 &  0.30&  1.24& 1.07 & 0.84 &  0.69\\
        \midrule
         cGan~\cite{chen2019unpaired} & unp. &49.87& 78.08&32.25&38.95 &44.91 & 48.81 &2.62 & 1.77  & \textbf{4.69} & 6.11&5.38 & 4.11 \\
        SInv~\cite{zhang2021unsupervised}& unp. & 54.87 & 78.68 & n/a &55.70 &62.31 &62.89 & 4.00 & \textbf{1.64} & n/a &\textbf{5.57} &\textbf{4.74} & \textbf{3.99}\\
        \midrule
        P2C~\cite{cui2023p2c} & ssp. & 39.43 & 68.56 & 38.09 & 48.74 & 57.67 & 50.50 & 7.54 & 3.37 & 13.45 & \underline{7.92} & \underline{5.21} & 7.50\\
         Ours & ssp.  & \textbf{68.63}&  \textbf{81.49} &  \textbf{53.73} &\textbf{56.02}  &\textbf{65.68} & \textbf{65.11}  & \textbf{1.75} & \underline{2.12} & \underline{9.58} &8.19  &11.39 & \underline{6.61} \\
         \bottomrule
    \end{tabular}
    \caption{Quantitative results of shape completion on our dataset. Type ``sp.'' and ``ssp.'' represent supervised and self-supervised methods, ``unp.'' implies the unpaired method. We write the best performance among all unpaired and self-supervised methods in bold and underline the better performance in self-supervised methods.}
    \label{tab:shapenet_vis}
\end{table}

\subsection{Implementation Details}
We use three layers of set abstraction (SA) modules and two point transformers to learn the shape code and generate 128 seeds of the missing part in the completion function $\mathcal{G}$. The seeds are further merged with the input partial points to create a coarse point set. From this point set, 512 points are extracted in total by farthest point sampling (FPS). The Upsampler is designed from ~\cite{xiang2021snowflakenet}, which outputs 2048 points. $\mathcal{T}$ contains a warp function $D$ with 10 layers of MLP and LSTM. The UDF decoder $T$ consists of 4 MLP layers and a hyperbolic tangent function as the final activation, followed by the absolute operator to predict the UDF. We use the Adam optimizer with $\beta_1=0.9$ and $\beta_2=0.999$. The initial learning rate is 5e-4 and the continuous decay rate is 0.5 every 500 epochs. The model is trained for 2500 epochs on 4 NVIDIA RTX GPUs with a batch size of 24. More network architecture and training details can be found in the supplementary. The code and dataset can be found in \href{https://github.com/lmy1001/Self-supervised-Shape-Completion}{https://github.com/lmy1001/Self-supervised-Shape-Completion}.

\subsection{Partial Shape UDF (PartialUDF) Dataset}
There exist few datasets that provide partial shapes with implicit representations. 3D-EPN~\cite{dai2017shape} proposes a 3D partial shape dataset containing a truncated signed distance field (TSDF) not satisfying our UDF representation. PatchComplete~\cite{rao2022patchcomplete} provides the same representation based on ScanNet dataset~\cite{dai2017scannet}. Most partial shape datasets such as MVP dataset~\cite{pan2021variational}, and PCN dataset~\cite{wang2018pcn} provide only the point cloud representation. These datasets typically offer a limited number of points (2048/4096/8192/16384), which are not enough to generate dense surface and UDF samples. To address this limitation, we develop a partial shape dataset containing dense partial surfaces and UDF samples for each instance.

\textbf{Dataset preparation.} We develop our dataset based on ShapenetCore.v2~\cite{chang2015shapenet} that provides a wide range of water-tight objects across different categories. Our PartialUDF-Shapenet dataset contains diverse categories (cars/lamps/sofas /planes/tables/chairs/bathtubs/beds). For each object instance, we carefully position 30 cameras in a uniform distribution on a unit sphere which creates 30 partial observations, similar to MVP dataset~\cite{pan2021variational}. For each partial shape, we densely sample 30,000 points on the seen surface and compute UDF around them following DeepSDF~\cite{park2019deepsdf}. We use the same train/test splits as in the DeepSDF~\cite{park2019deepsdf}. For each object instance, we randomly select 6 partial observations to collect the training set, in order to make the training set large. During inference time, we randomly test 2 partial inputs for each instance. Note that, we make sure that each mini-batch during training/inference does not contain the partial observation from the same shape instance, therefore, avoiding multi-view knowledge during training and test. Furthermore, we extend the PartialUDF dataset to the non-rigid DFaust~\cite{bogo2017dynamic} dataset (PartialUDF-DFaust) which provides the ground-truth (GT) correspondences. We use 7 sequences for training set and 2 sequences for test set. More details are in the supplementary.

   \begin{table}[t]
    \begin{minipage}{0.48\textwidth}
   \centering
    \fontsize{8}{10}\selectfont
    \setlength{\tabcolsep}{1.5mm}
    \begin{tabular}{l| c | c c c}
 \toprule
         Methods & type & F1 & CD & Corr $\ell$2\\
         \midrule
         SeedF~\cite{zhou2022seedformer} & sp. & 81.18  & 2.82& - \\
         cGan~\cite{chen2019unpaired} & unp. & 57.90  & 7.34 & -  \\
         \midrule
         P2C~\cite{cui2023p2c} & ssp. & 42.27 & 7.50 & - \\
         INR\_only & ssp. & 80.29 & 2.84 & 0.34\\
         Ours & ssp. &
         \textbf{84.57} & \textbf{1.76} & \textbf{0.28} \\
          \bottomrule
    \end{tabular}
    \caption{Quantitative results on DFaust dataset.
    Our method performs the best.}
    \label{tab:quantitave_on_dfaust.}
    \end{minipage}
    \begin{minipage}{0.48\textwidth}
    \centering
    \fontsize{8}{10}\selectfont
    \setlength{\tabcolsep}{2.3mm}
     \begin{tabular}{l|c |  c  c }
     \toprule
        Methods & type & Fidelity & MMD \\
        \midrule
         SeedF~\cite{zhou2022seedformer} & sp. &  2.11 &1.22 \\
         GRNet~\cite{xie2020grnet} & sp. & 12.97 & 1.17\\
         SInv~\cite{zhang2021unsupervised} & unp. & \textbf{1.88} & 1.11 \\
         \midrule
         P2C~\cite{cui2023p2c} & ssp. & 15.32 & 1.20 \\
         Ours & ssp. & 4.61 & \textbf{0.08} \\
         \bottomrule
    \end{tabular}
    \caption{Quantitative results on KITTI. We achieve the best MMD metrics.}
    \label{tab:kitti_test}
    \end{minipage}
   \end{table}

\subsection{Shape Completion on PartialUDF-Shapenet}
\textbf{Quantitative results.} We evaluate the performance of our method on the PartialUDF-Shapenet dataset across cars, planes, sofas, chairs, and tables categories. We compare our method with the supervised SeedFormer~\cite{zhou2022seedformer}, the unpaired cGan~\cite{chen2019unpaired}, the pre-trained GAN-based method ShapeInversion~\cite{zhang2021unsupervised}, and the self-supervised method P2C\cite{cui2023p2c}. We train SeedFormer, cGan, and P2C models on our train set and directly use the provided trained model from ShapeInversion and test on our test set. We measure F1 score ($\tau=0.03$) and CD errors, results are displayed in Table~\ref{tab:shapenet_vis}. 
Although our method cannot compete with the supervised SeedFormer, we get the second-best F1 score in all categories, and the second-lowest CD error in the cars category. Moreover, we beat the self-supervised P2C in most categories. However, it is worth noting that for complex categories like chairs and tables, our method performs worse in CD errors. Our method has no prior on whether the completed shape contains additional parts (\eg, table compartments), thus it tends to complete shapes aggressively with more parts than that exist in the shape instance. 

\begin{figure}[t]
 \begin{minipage}{0.52\textwidth}
    \centering
    \includegraphics[width=1.0\textwidth]{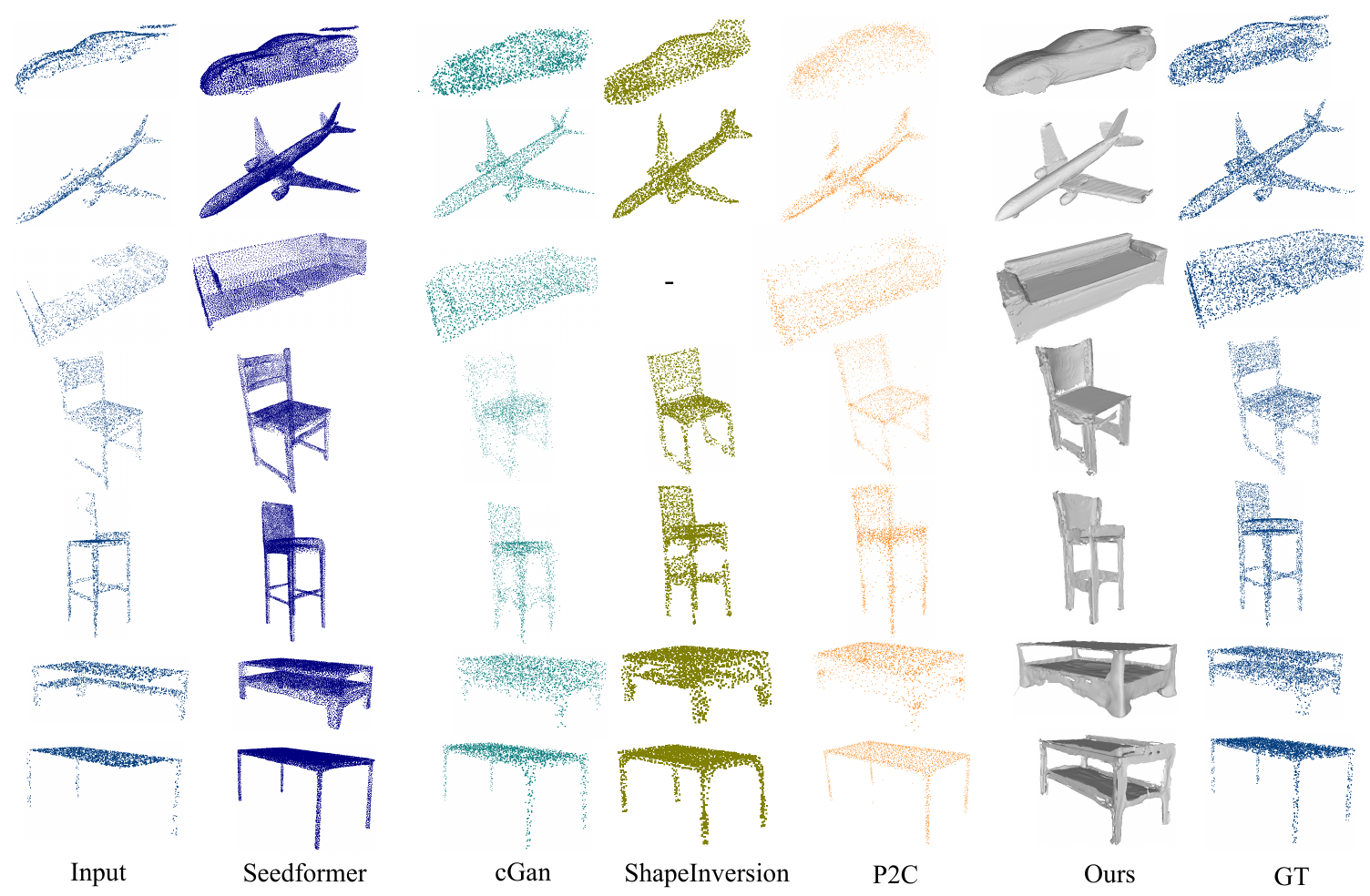}
    \caption{Qualitative results of shape completion on Shapenet various categories, compared with SeedFormer~\cite{zhou2022seedformer}, cGan~\cite{chen2019unpaired}, ShapeInversion~\cite{zhang2021unsupervised}, P2C~\cite{cui2023p2c}, our method completes dense meshes.}
    \label{fig:shapenet_vis}
    \end{minipage}
    \begin{minipage}{0.45\textwidth}
    \centering
        \includegraphics[width=1.0\textwidth]{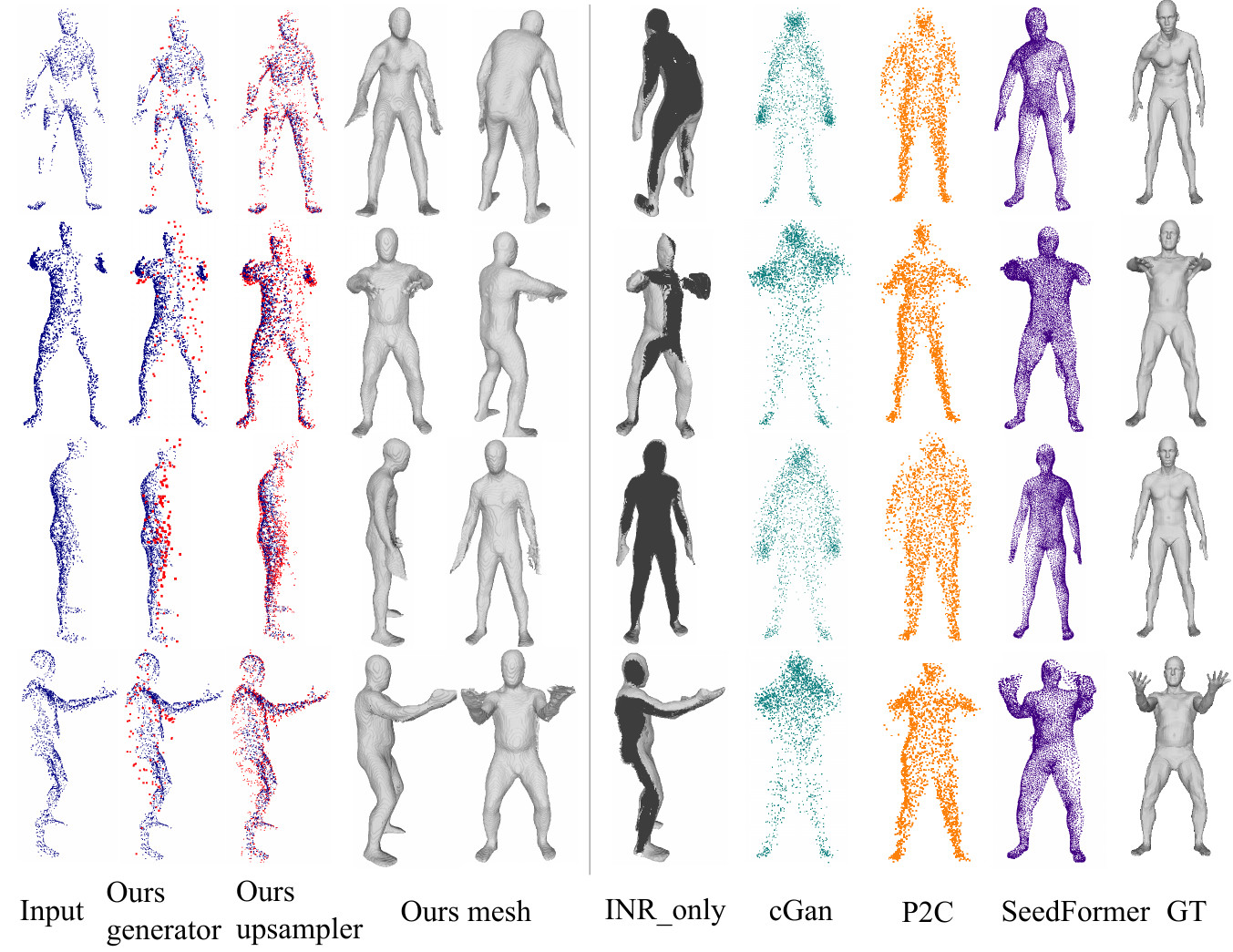}
    \caption{Qualitative results on DFaust. We show generated points (in red) from the generator (completion function) and the upsampler, and meshes compared with other methods.}
    \label{fig:dfaust_comparison}
    \end{minipage}
\end{figure}

\textbf{Qualitative results.} We present some qualitative results in Figure~\ref{fig:shapenet_vis}. Given the partial observations, our method shows impressive completion performance. cGan~\cite{chen2019unpaired} always completes noisy shapes and ShapeInversion~\cite{zhang2021unsupervised} sometimes fails to keep local input parts. P2C~\cite{cui2023p2c} completion is noisy and fails sometimes, especially when there is large missing part in the input shape, see the first three rows in Figure~\ref{fig:shapenet_vis}, P2C almost fails to generate the part while our method complete the mesh. Conversely, SeedFormer generates detailed shapes closely resembling the ground truth. Our method generates dense meshes and preserves the input points by design. Note that, due to the use of a common template space, our method may generate non-existent but valid parts, as seen from the last row in Figure~\ref{fig:shapenet_vis}. The table instance without the lower compartment generates it while completing the missing leg. 
This behavior results in worse CD errors. 

\subsection{Shape Completion and Correspondences on PartialUDF-DFaust}
\textbf{Quantitative results. } We train our method on the non-rigid dataset and test its performance. We compare the method with the supervised SeedFormer~\cite{zhou2022seedformer}, unpaired cGan~\cite{chen2019unpaired} and self-supervised P2C~\cite{cui2023p2c} methods. Moreover, we additionally compare with the INR\_only method which consists of only the template-based INR module. Meanwhile, we measure the ``Corr $\ell2$'' error that denotes the difference between the predicted correspondences (by finding the nearest neighbor in the template filed) and the GT correspondences provided by the DFaust dataset~\cite{bogo2017dynamic} following the protocol in \cite{wang2019dynamic}. Results are presented in
Table~\ref{tab:quantitave_on_dfaust.}. It shows great improvement with our method in both shape completion and dense correspondences. More importantly, the PartialUDF-DFaust dataset has an average missing rate of around $72\%$, and our method still achieves a high F1-score. We are 4 times smaller in CD errors compared with the supervised SeedFormer~\cite{zhou2022seedformer}. At the same time, we also improve the dense correspondences compared with INR\_only by gradually completing the template space, which proves that the completion module is also helping the template-based INR module to learn the template shape well. One reason for the excellent performance on DFaust~\cite{bogo2017dynamic} is that the common template space exactly satisfies the problem at hand. Thus together with the involution constraint, the completion function quickly learns to fill in the missing parts correctly. P2C, however, performs even worse than INR\_only due to its weak ability in non-rigid shapes and the large missing rates. More analysis of correspondences is displayed in supplementary.

\textbf{Qualitative results. } We present qualitative results in Figure~\ref{fig:dfaust_comparison}. The figure showcases 4 partial inputs with various gestures, alongside the completion process of our method, INR\_only, cGan~\cite{chen2019unpaired},  P2C~\cite{cui2023p2c}, SeedFormer~\cite{zhou2022seedformer}, and GT meshes respectively. Our method completes the water-tight meshes that resemble the ground-truth meshes faithfully. While cGan, P2C and SeedFormer fail to deal with large non-rigid deformations, \eg, the arms. Furthermore, we showcase the generated template and the dense correspondences in Figure~\ref{fig:dfaust_correspondences}. The correspondences are presented across the same colors. Our method generates the template shape more like a mean shape of the category with good correspondences while the INR\_only method results in the wrong correspondences.

   \begin{figure}[t]
    \begin{minipage}{0.60\textwidth}
\centering
\includegraphics[width=0.95\textwidth]{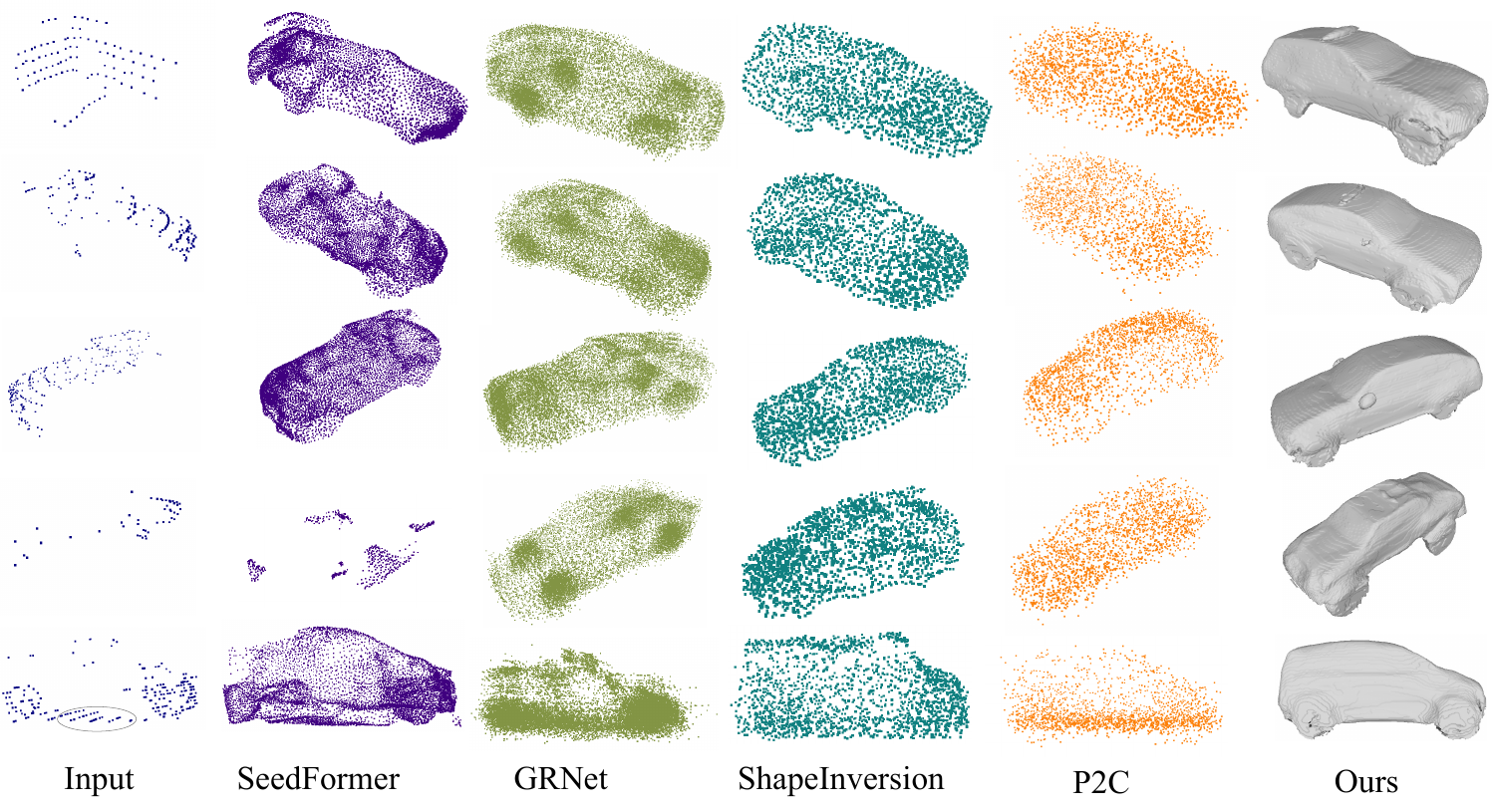}
    \caption{Qualitative results on KITTI dataset. Our method completes the partial data even when there are extremely few points available. }
    \label{fig:kitti_comparison}
\end{minipage}
        \begin{minipage}{0.36\textwidth}
            \centering
        \includegraphics[width=1.0\textwidth]{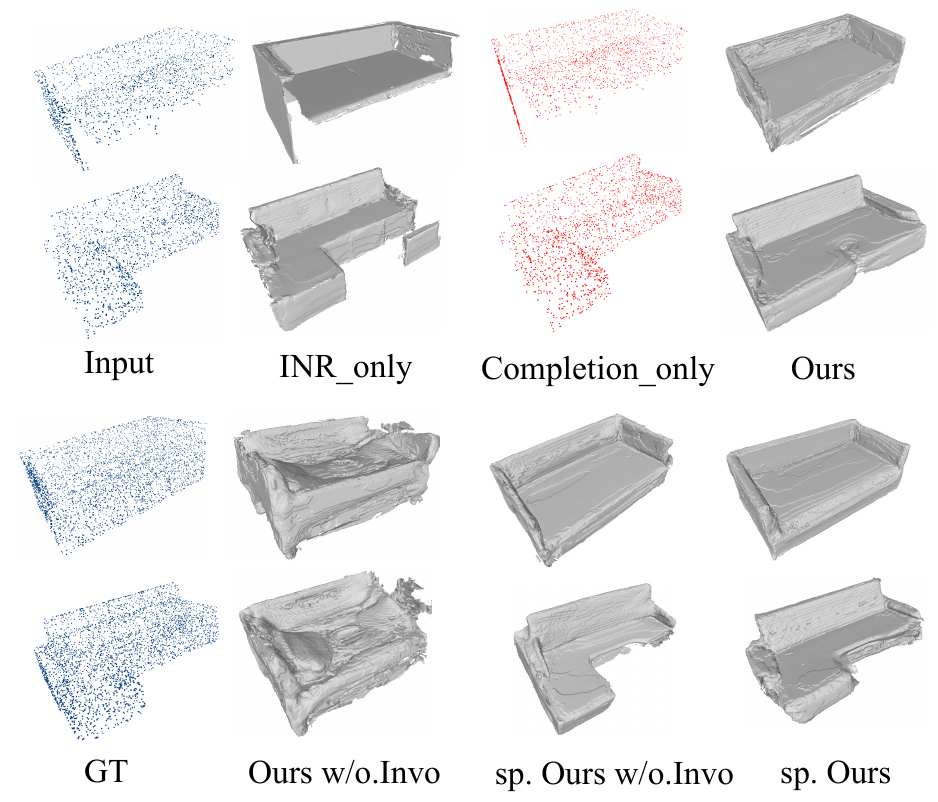}
        \caption{Visualization results of ablation study on sub-modules with different settings.}
        \label{fig:ablate_on_modules}
        \end{minipage}
    \end{figure}

\subsection{Shape Completion on Real-World Scans}
\textbf{Quantitative results.} We investigate the completion capability further in real-world scans extracted from KITTI~\cite{geiger2013vision} dataset. Compared with the synthesized partial datasets, real-world data is noisier and sparser, increasing the difficulty of completion. For the test on KITTI dataset, we follow the protocol of GRNet~\cite{xie2020grnet} to prepare the test data. For all the methods, we use the trained models from the ShapeNet cars category and test the performance on KITTI. We evaluate our method following SeedFormer~\cite{zhou2022seedformer} with two metrics: Fidelity Distance, which measures the distance from every partial scan to its completed shape, denoting how well the input is reserved. The other metric is the Minimal Matching Distance (MMD), measuring the Chamfer Distance between the completed shape to its closest shape in the ShapeNet cars category, it represents the similarity between the completed shape to the synthetic shape.  Results are presented in Table~\ref{tab:kitti_test}. Although our method is slightly worse in Fidelity, we achieve the lowest MMD, even better than the supervised method, showing that our method generates more car-like shapes. 

\textbf{Qualitative results.} We further present the qualitative results on KITTI scans in Figure~\ref{fig:kitti_comparison}. Our method generates the complete car even with extremely few points while SeedFormer fails and GRNet raises many noisy points, P2C on the other hand, always generates fewer points in the missing area. Moreover, when there are outliers in the input scans, \eg, the points circled out in the last row (cars should not contain points that are on the ground), while ShapeInversion and P2C still preserve the points, our method ignores the outliers, which results in slightly worse Fidelity.

\subsection{Ablation Study }
We analyze the effectiveness of each module of our network with different setups. First of all, we test the effectiveness of the completion module by training the template-based INR module only (INR\_only) and vice versa (Completion\_only). We then measure the capability of involution loss by training the network without self-involution (Ours~w/o. Invo). We further train the network in a supervised way (sp.~Ours) by supervising the completion module with the complete point clouds instead of the learned template shape, and the supervised way without involution loss (sp.~Ours~w/o.~Invo). All the experiments are trained and tested on the sofas category. Results are presented in Table~\ref{tab:ablate_modules}. More visualizations are presented in Figure~\ref{fig:ablate_on_modules}.

\textbf{Effet of Completion module.} Compared with INR\_only, the CD error decreases by 61\% after adding the completion module. Moreover, see in Table~\ref{tab:quantitave_on_dfaust.}, the correspondences are improved, meaning that the generated parts are helping with the build of the consistent template shape. 

\textbf{Effect of template-based INR module.} In Figure~\ref{fig:ablate_on_modules}, Completion\_only generates only the input shape.  we show the great importance of the template-based INR module providing the supervision on the completion module.

\textbf{Effect of involution.} In the self-supervised setting, the involution constraint plays an essential role in training the completion function. The network performs even worse than INR\_only if trained without involution loss. From Figure~\ref{fig:ablate_on_modules}, we find the method without involution loss tends to complete the same noisy mesh in the field. Furthermore, in the supervised setting, involution is still beneficial in decreasing CD errors, as it encourages the completion function towards completing the missing parts only. 

\textbf{Compared with supervised setting.} We evaluate our method in a supervised setting. We observe an obvious decrease in CD errors, as well as an improvement in the F1 score. The supervised setting is powerful in completing accurate point clouds and thus is beneficial in generating the dense complete mesh, resulting in both lower CD errors and better F1 scores.
    \begin{table}[t]
        \centering
    \fontsize{8}{10}\selectfont
    \setlength{\tabcolsep}{1.8mm}
     \begin{tabular}{l| c | c  c }
     \toprule
        Methods & type & F1 & CD \\
        \midrule
        INR\_only & ssp. & 51.54 & 24.52 \\
        Completion\_only & ssp. & 34.72 & 15.90 \\
        Ours & ssp. & \textbf{53.73} & \textbf{9.58} \\
        \bottomrule
    \end{tabular}
    \begin{tabular}{l| c | c  c }
        \toprule Methods & type & F1 & CD \\
        \midrule
        Ours w/o. Invo & ssp. & 16.31 & 24.01 \\
        Ours & ssp. & \underline{53.73} & \underline{9.58} \\
        \bottomrule
        sp.~Ours~w/o. Invo & sp. &\textbf{55.58} & 7.19 \\
        sp.~Ours & sp. & 55.31 & \textbf{6.67} \\
         \bottomrule
    \end{tabular}
    \caption{Ablation on sub-modules. We ablate the effectiveness of each module by separately training each part. All ablation studies are performed in the sofas category.}
    \label{tab:ablate_modules}
    \end{table}

\begin{figure}
\begin{minipage}{0.60\textwidth}
\small
\fontsize{8}{10}\selectfont
\setlength{\tabcolsep}{0.5mm}
    \centering
    \begin{tabular}{c| c c c c | c c c c}
    \toprule
           \multirow{3}{*}{views} & \multicolumn{4}{c|}{F1} & \multicolumn{4}{c}{CD} \\
          & SeedF & cGan & P2C & Ours & SeedF & cGan & P2C & Ours \\
           & sp. & unp. & ssp. & ssp. & sp. & unp. & ssp. & ssp. \\ 
         \midrule
         1 & \textbf{76.51} & 47.53&38.19 &  \underline{56.39} & \textbf{2.30} & 3.93 &6.61 &  \underline{2.59}\\
         2 & \textbf{78.53} & 49.16&37.44 & \underline{62.29} & \underline{2.20} & 4.25 & 6.29 &  \textbf{2.12} \\
         4 & \textbf{81.52} & 50.84 &37.34 & \underline{66.63} &\underline{2.07} & 3.68 & 6.42 & \textbf{1.73} \\
         6 & \textbf{80.85} & 49.87&39.43 & \underline{68.63} & \textbf{0.82} & 2.62 & 7.54 & \underline{1.63}\\
         \midrule
         1 & \textbf{68.17} & 34.90 &27.16 &\underline{41.18} & \textbf{3.07} &\underline{6.32} &15.48  & 14.23 \\
         6 &  \textbf{71.30} & 32.25 & 38.09 & \underline{53.73} & \textbf{1.24} & \underline{4.69} & 13.45 & 9.58 \\
      \bottomrule
    \end{tabular}
    \caption{Quantitative results of shape completion in cars (the above sub-table) and sofas (the below sub-table) categories. The training data contains 1/2/4/6 views of partial shapes for each instance respectively. The best performance is shown bold, and the second best performance is shown underlined.}
    \label{tab:views}
\end{minipage}
\begin{minipage}{0.38\textwidth}
    \centering
        \includegraphics[width=0.90\textwidth]{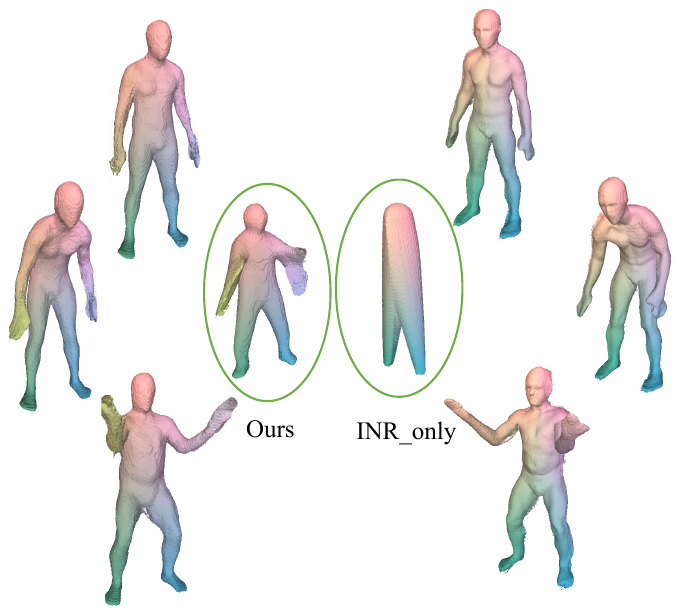}
        \caption{Visualization of dense correspondences and the generated template shape from our methods and the INR\_only method. Points with the same color are corresponding points.}
        \label{fig:dfaust_correspondences}
\end{minipage}
\end{figure}

\textbf{Train with smaller dataset. } Although most partial shape dataset contains more than one partial observation for each instance, for example, the popular PCN~\cite{wang2018pcn} provides 8 random views, 3D-EPN~\cite{dai2017shape} uses 6 views, and we know P2C can work with only one partial observation for each instance. Therefore, we experiment our method with much smaller dataset, consisting of 1/2/4/6 partial observations respectively, and test on the same test set. The results are presented in Table~\ref{tab:views}. We test on cars and sofas categories. Note that, 6 views are the common settings we use for all the methods. Our method still performs better than cGan and P2C with a single view on cars, as the involution and consistent correspondences are relatively strong priors. And all methods are not performing well having only one partial observation in each instance.

\section{Conclusions}
In this paper, we presented an approach for partial shape completion based on the geometric prior of deformation warp consistency and the involution self-loss of completion. We verified two important hypotheses. First, correspondences or a common template space can be used as a supervisory signal for the completion task. Second, a completion function can be formulated as an involution, if it is constructed to predict the missing part. Our experiments showed that both contributions can play important roles in guiding the completion module. Surprisingly, in some cases such as the human body completion, our method achieves impressive performance. Similarly, we observe that our method, despite not using any complete examples, out-competes several unpaired completion approaches in the F1 score, in the rigid category instances.



\section*{Acknowledgements}
The research is partially funded by VIVO collaboration project on real-time scene reconstruction.

%
%

\end{document}